\documentclass[lettersize,journal]{IEEEtran}
\usepackage{amsmath,amsfonts}
\usepackage{algorithmic}
\usepackage{algorithm}
\usepackage{array}
\usepackage[caption=false,font=normalsize,labelfont=sf,textfont=sf]{subfig}
\usepackage{amssymb}
\usepackage{textcomp}
\usepackage{booktabs}
\usepackage{stfloats}
\usepackage{url}
\usepackage{verbatim}
\usepackage{graphicx}
\usepackage{cite}
\usepackage{xspace}
\usepackage{xcolor}
\usepackage{multirow}

\usepackage[normalem]{ulem}

\newcommand{\etal}{\textit{et al.}}
\newcommand{\ie}{\textit{i.e.}}
\newcommand{\eg}{\textit{e.g.}, }

\begin{document}

\title{Focus Where It Counts: A Salience-Driven Vision-Language Model for Low Vision Assistance}

\author{
Jiazhao Liang$^{\dagger}$, Hao Huang$^{\dagger}$, Shuaihang Yuan$^{\dagger}$, Congcong Wen,
Geeta Chandra Raju Bethala,\\
Giles Hamilton-Fletcher, Yu Hao, John-Ross Rizzo, Mengyu Wang, Anthony Tzes, and Yi Fang%
\thanks{\textit{$^\dagger$Equal contribution.}}%
\thanks{Jiazhao Liang is with New York University Tandon School of Engineering, Brooklyn, NY 11201, USA.}%
\thanks{Hao Huang, Yu Hao, Geeta Chandra Raju Bethala, Congcong Wen, Shuaihang Yuan, Anthony Tzes, and Yi Fang are with NYUAD Center for Artificial Intelligence and Robotics (CAIR) and Embodied AI and Robotics (AIR) Lab, New York University Abu Dhabi, Abu Dhabi 129188, UAE.}%
\thanks{Giles Hamilton-Fletcher and John-Ross Rizzo are with NYU Grossman School of Medicine, NYU Langone Health, New York, NY 10016, USA.}%
\thanks{Mengyu Wang is with Harvard AI and Robotics Lab, Harvard University, Boston, MA 02114, USA.}%
}

\maketitle

\begin{abstract}
Vision-language models (VLMs) are rapidly progressing and offer promising capabilities for assistive technologies supporting persons with blindness or low vision. {However, existing VLMs are primarily designed for general-purpose captioning and do not explicitly model human perceptual priorities, thereby limiting their ability to emphasize the most relevant information in a scene.} To address this gap, we propose a salience-driven captioning framework that prioritizes scene elements according to their importance for human-centered assistance.  {We curate three salience-aware datasets, namely, Salience COCO, Salience Flickr, and Salience VizWiz, with object-level salience annotations designed to reflect the visual information most relevant to low vision users across different environments. }{Building on these datasets, we introduce Salience-LLaVA, a salience-aware VLM that incorporates salience cues to generate captions in which important elements are mentioned in the order of importance. Our work makes four main contributions. We build salience-aware datasets verified by low vision participants, propose Salience-LLaVA to describe objects in the order of importance, introduce SCMI to evaluate ordering accuracy, and deploy the system on assistive glasses to demonstrate real-world practicality.} Code and datasets are available at: https://github.com/topo-focus/Topofocus
\end{abstract}

\begin{IEEEkeywords}
Assistive technology, computer vision, image captioning, vision-language models
\end{IEEEkeywords}

\section{Introduction}

\IEEEPARstart{R}{ecent} studies show that vision-language models (VLMs) can describe images in a zero-shot manner, enabled by training on large and diverse datasets. This generalization has supported their use in applications such as visual interpreters that convert complex scenes into natural language~\cite{guo2023images,wen2025zero}. More recently, VLMs have been integrated into wearable devices~\cite{hao2024chatmap}, offering real-time feedback for individuals with low vision. By translating visual input into verbal descriptions, these systems help users understand and navigate dynamic environments~\cite{baig2024ai}. {However, current VLMs are built for general-purpose captioning and describe scenes without considering how humans naturally prioritize visual information, often missing elements critical to the safety and orientation of low vision users.}

\begin{figure}[t]
    \centering
    \includegraphics[width=0.90\columnwidth]{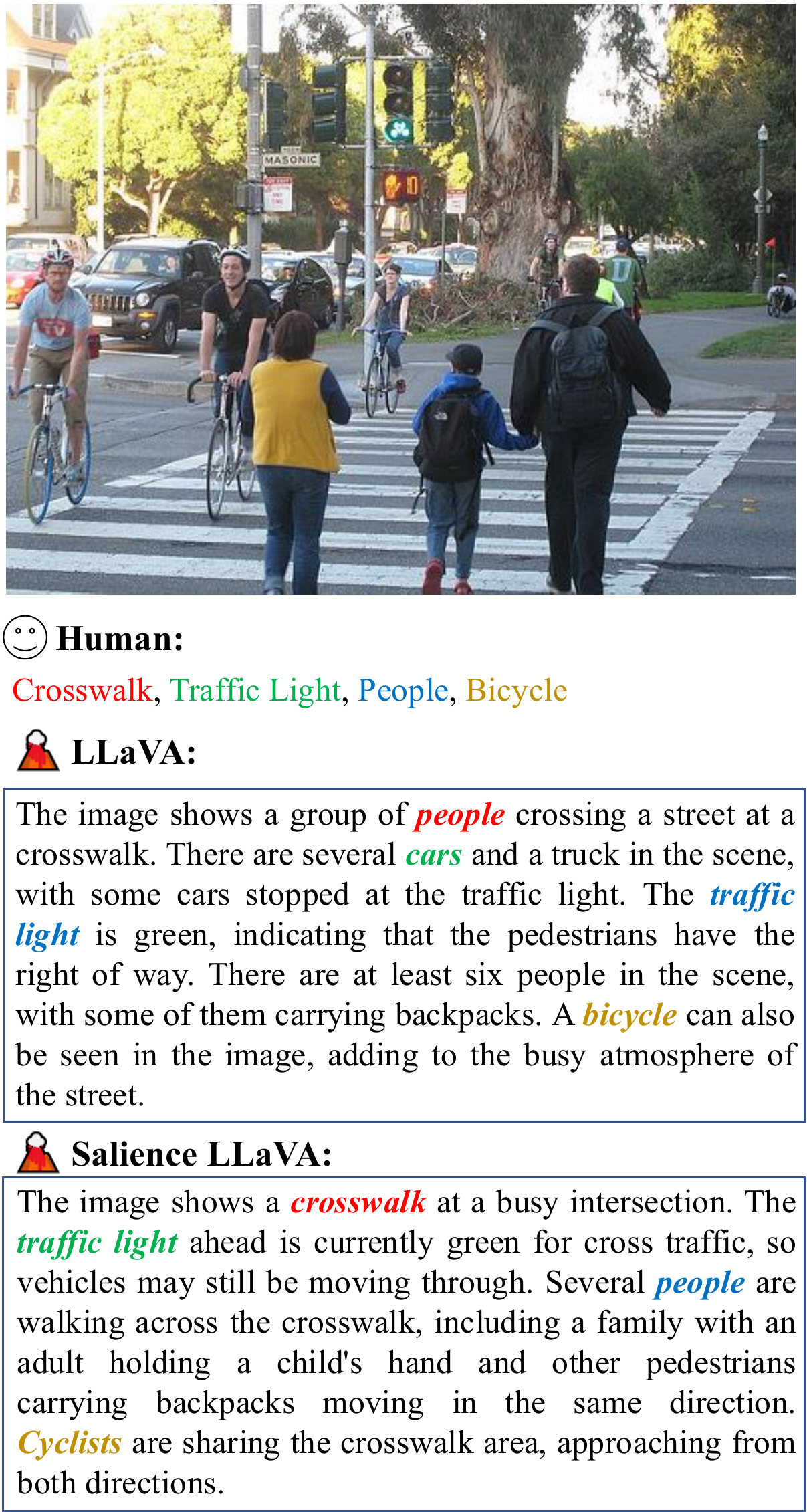}
    \caption{Comparison of image descriptions generated by the original LLaVA and Salient-LLaVA. The first row lists salient objects in order of importance, as determined by a human annotator, representing human commonsense prioritization. While the original LLaVA description does not fully align with human perception, Salient-LLaVA better reflects human-centered relevance by highlighting key elements earlier in the caption.}
    \label{fig:intro}
\end{figure}

This limitation becomes critical in assistive settings, where task-relevant and safety-critical cues are essential for independent mobility. Although VLMs produce detailed outputs, they often miss subtle but important environmental cues. These include uneven curbs, oncoming vehicles, or misaligned tactile paths. {As shown in Fig.~\ref{fig:intro}, a human annotator ranks salient objects in a street-crossing scene in the descending order of importance: \textit{Crosswalk}, \textit{Traffic Light}, \textit{People}, and \textit{Bicycle}. The original LLaVA caption begins with \textit{People} rather than \textit{Crosswalk} and places \textit{Vehicles} before \textit{Bicycle}, deviating from this human priority. Salience-LLaVA more closely follows the human ordering, describing the \textit{Crosswalk} and \textit{Traffic Light} first before mentioning \textit{People} and \textit{Cyclists}. These challenges highlight the need for VLMs that can prioritize scene elements based on their importance to the user, rather than treating all objects equally in the description.}

{In practice, visually impaired users often have a low visual acuity, a limited field of view, or they experience night blindness, making small but critical objects easy to miss~\cite{reynolds2024salient, tavakoli2017paying}. For these users, the order in which scene elements are described matters significantly. Safety-critical information, such as road conditions, traffic signals, and moving obstacles, must be conveyed first, as delayed or buried descriptions may lead to dangerous decisions. A caption that accurately lists all objects in a scene but fails to prioritize them according to the user's immediate needs provides limited practical value. Therefore, there is a clear need for a salience-aware captioning mechanism that orders scene elements based on their importance to the user's safety and navigation.}

{This gap stems from the absence of any mechanism to model 
human perceptual priorities: standard VLM captioning pipelines 
assign no ordering or weight to scene elements based on user needs. For example, although a sighted person may naturally disregard certain obstacles on the ground, failing to account for them can present substantial trip hazards for individuals with visual impairments. Recent work in navigation and anomaly detection~\cite{liang2024visarl, son2025infrastructure, liu2025situat3dchange} also underscores the importance of salience tuned for people with visual impairments. However, despite these advances, existing captioning models lack mechanisms to incorporate salience information that reflects human perceptual priorities. Human-centered alignment is critical in assistive captioning because low vision users may rely entirely on verbalized descriptions to make safety-critical decisions, making the ordering of information as important as its content. Inspired by saliency maps, we propose a saliency-aware captioning system that not only describes scene elements but also prioritizes and orders them based on their contextual importance under specific circumstances. We approach this problem from both the \textit{dataset curation} and \textit{model architecture} perspectives by curating salience-annotated datasets grounded in the real needs of users with low vision and by explicitly integrating saliency rankings generated by a dedicated saliency prediction module into the captioning pipeline, ensuring that the most relevant elements are described first.} Our main contributions are summarized as follows.

\begin{enumerate}
    \item \textit{Datasets With Salience-Aware Captions Tailored to Low Vision Users:} We present Salience COCO, Salience Flickr, and Salience VizWiz to capture the complex, real-world human-centered salient objects missing from existing image captioning benchmarks.
    
    \item \textit{Salience-Aware VLM for Captioning:} We integrate image salience features into a VLM to generate descriptions that prioritize visually crucial details often overlooked by generic VLMs.
        
    \item \textit{Assistive Glasses Implementation for Low Vision Users:} We integrate our salience-aware captioning model with glasses hardware that capture visual input and deliver audio feedback, offering an accessible and intuitive experience for the visually impaired.
\end{enumerate}

By integrating salience information with a VLM-based captioning model, our approach aims to bridge the gap between generic, visually prominent descriptions and the specific, human-centered requirements of low vision users. We believe that our work will advance the development of inclusive, reliable, and user-centered assistive technologies.

\section{Related Works}

\subsection{VLM for Image Captioning}
VLMs have advanced image captioning by effectively bridging visual inputs with natural language outputs through diverse architectural innovations~\cite{lin2024rs,lin2025fedrsclip}. For instance, VinVL~\cite{zhang2021vinvl} employs object detection to extract detailed visual features that significantly boost caption accuracy. Building on this, SimVLM~\cite{wang2022simvlm} enhances captioning performance by utilizing a straightforward prefix language modeling objective over vast, weakly supervised image-text pairs. Complementing these approaches, both OFA~\cite{wang2022ofa} and BLIP~\cite{li2022blip,li2023blip} adopt unified sequence-to-sequence frameworks to streamline caption generation. In addition, mPLUG~\cite{li2022mplug} integrates cross-modal skip connections to better balance visual and linguistic cues, thereby reducing issues such as hallucinations. ViTCAP~\cite{fang2022injecting} elevates caption fidelity by incorporating semantic concept tokens within a detector-free Transformer~\cite{vaswani2017attention} to enrich the descriptive quality of generated captions. Meanwhile, reducing the cost of data annotation remains a key concern, as active learning techniques have shown promise in efficiently curating labeled datasets for visual recognition tasks~\cite{amin2023deep}. {Beyond captioning, recent MLLM-based methods explore federated adaptation and embedding-level reranking~\cite{xu2024fedmllm, gu2026unime}, while augmentation-based methods strengthen contrastive representations for recognition tasks~\cite{xu2026attack, liu2026rar}; however, these directions emphasize model training, candidate discrimination, or recognition robustness rather than human-centered object prioritization in captioning.} Despite their strong performance, these methods share a common limitation: they treat all visual elements with equal importance during caption generation, without mechanisms to prioritize objects based on their relevance to the user. On the contrary, our work integrates salience cues into the captioning pipeline, enabling the model to emphasize safety-critical objects in the descriptions.

\subsection{Salience Information as Human-Centered Cues}
Salience maps are designed to highlight the most influential regions or features in an input image, aligning model focus with human visual priorities. For instance, Chen \etal~\cite{chen2023deep} demonstrate that deep saliency models can decompose learned representations into interpretable semantic bases that mirror human fixation patterns. Similarly, Levin~\etal \cite{levin2022models} propose parameter-space saliency maps to diagnose and correct misalignments between model attention and human perception. Furthermore, Deng \etal~\cite{deng2024advancing} leverage real human fixation data to accurately rank salient regions in images, thereby enhancing the reliability of saliency maps. In parallel, Bhunia \etal~\cite{bhunia2023sketch2saliency} show that human sketches can serve as effective weak supervision for saliency detection, emphasizing the value of natural perceptual cues. 
{Moreover, Chen \etal~\cite{chen2024gazexplain} generate natural language explanations for visual scan paths, linking model decisions with interpretability. Related work on video anomaly detection also highlights the need to identify critical visual elements in dynamic scenes~\cite{ul2024video,ul2022eadn}. However, these methods mainly interpret attention patterns rather than guide language generation. Our work incorporates saliency rankings into a VLM captioning framework, so that generated descriptions follow the importance ordering aligned with low vision users.

}

\section{Datasets Curation}
\begin{figure*}[htbp]
    \centering
    \includegraphics[width=\textwidth]{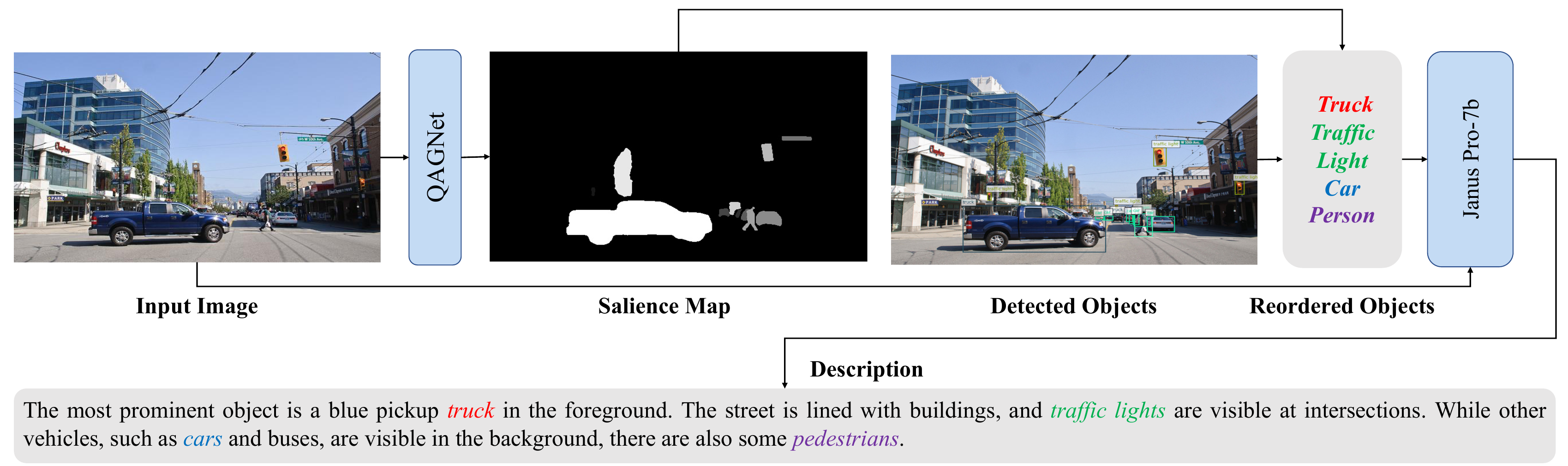}
    \caption{An Illustration of the salience-aware captioning ground-truth generation process. The input image is processed to produce a salience map, from which salient objects are detected and reordered. A salience-aware description is then generated, highlighting the most salient elements in the scene.}
    \label{fig:dataset}
\end{figure*}
Recent advances in vision-language research have significantly enhanced assistive devices for low vision individuals; however, the models powering these technologies are trained on datasets that suffer from several limitations. First, the original captions do not capture \textit{salient} information; in other words, they fail to convey the relative importance of visual elements as humans perceive them. Second, some scenarios presented in the pictures lack the \textit{complexity} to reflect real-world environments. Third, the datasets are not designed with \textit{low vision} user experiences in mind, often including sporting scenarios that are nonadaptive, and hence less representative of their daily activities. These limitations underscore the need for salience-aware captioning datasets coupled with visually impaired user experiences, prompting us to repurpose and propose three datasets: Salience COCO, Salience Flickr, and Salience Vizwiz. {Our filtering criteria are grounded in both low vision research and empirical evidence from observed accessibility needs in real-world settings. Visual crowding is widely recognized as a major bottleneck in degraded and peripheral vision because it affects not only object detection but also object discrimination in cluttered scenes~\cite{whitney2011visual}. Many people with low vision rely on peripheral or residual central vision due to conditions such as macular degeneration. As a result, they often face greater difficulty in scenes that contain multiple object types and richer semantic content. These environments require users to distinguish which objects matter most, not just detect whether an object is present. This challenge is common in everyday settings such as kitchens, living rooms, and urban streets. On the contrary, images with only a single supercategory contain less perceptual ambiguity and lower semantic complexity. They therefore do not reflect the complexity of many real-world assistive situations. By requiring at least two supercategories, we ensure that the benchmark evaluates models under more realistic levels of visual complexity and better captures the perceptual demands faced by low vision users.}

 {We exclude outdoor sports scenes because prior evidence suggests that assistive technology for blind and low vision users is needed mainly in indoor everyday settings. Brady \etal~\cite{brady2013visual} analyzed more than 40,000 visual questions submitted by 5,329 blind users through the VizWiz application. Most requests focused on object identification, scene description, and text reading. These questions were concentrated in household environments, retail spaces, and indoor navigation tasks. Gurari \etal~\cite{gurari2018vizwiz} further showed that images captured by blind users mainly depict close-range indoor objects, personal items, and daily living scenes, with very limited representation of sporting events. Taken together, these studies indicate that outdoor sports scenes are not a common setting in which assistive visual support is typically sought.}

 {This choice is also consistent with prior evidence on everyday visual function and accessibility needs. Owsley \etal~\cite{owsley2007effect} showed that quality of life among older adults with impaired vision is closely tied to activities such as reading, self-care, and navigating residential environments. Outdoor sports scenes in COCO usually contain wide spaces, distant people, and fast motion. They are, therefore, visually and functionally different from the cluttered everyday environments that more often motivate assistive use. Including them would reduce the alignment between the benchmark and the practical situations it is intended to represent. Excluding them helps keep the dataset focused on scenes that better reflect the documented daily needs of low vision users.}

\begin{figure*}[t]
    \centering
    \includegraphics[width=0.99\linewidth]{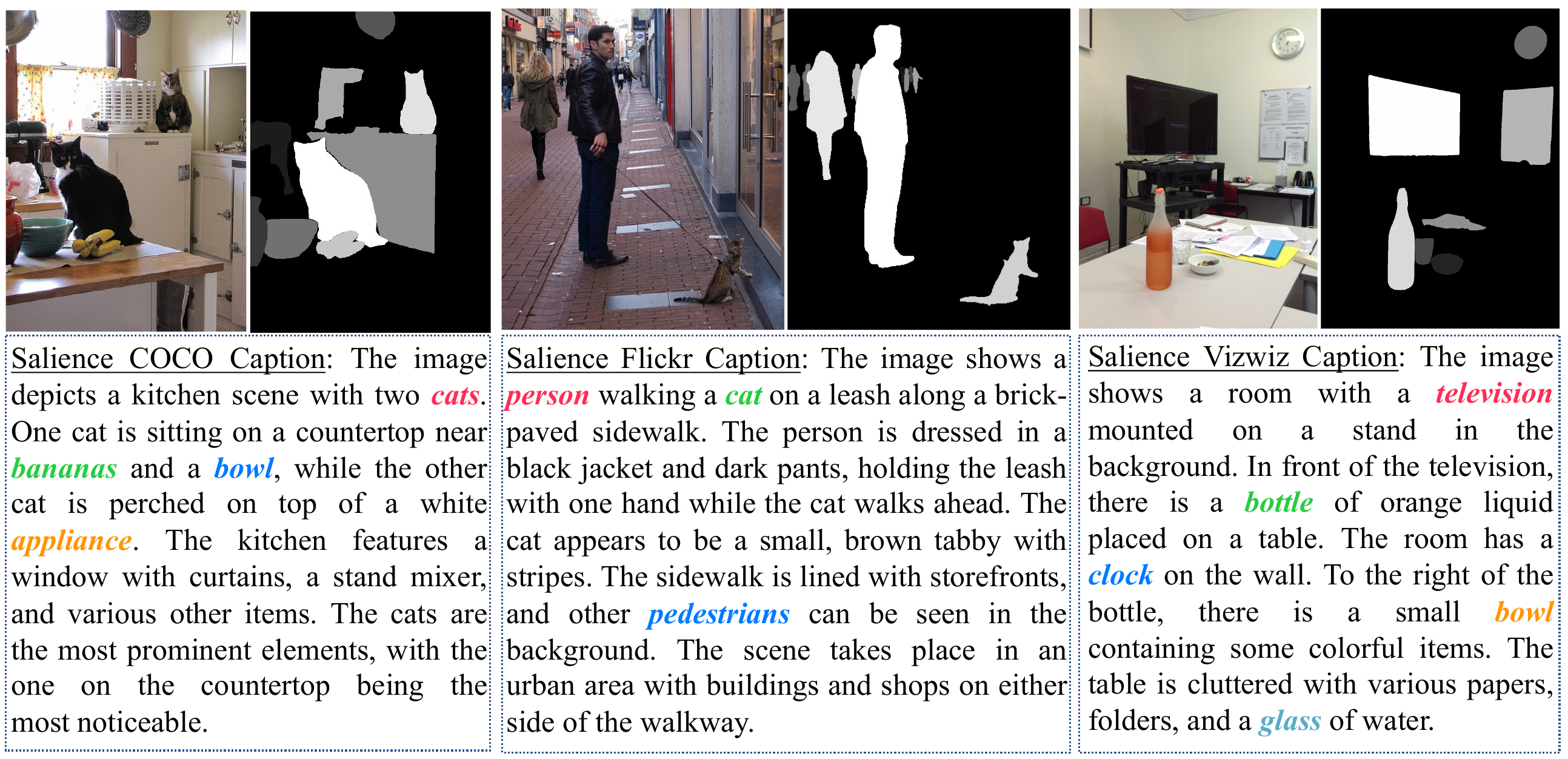}
    \caption{Examples of images, salience masks, and salience-aware ground-truth captions of Salience COCO, Salience Flickr, and Salience Vizwiz datasets.}
    \label{fig:data_demo}
\end{figure*}

\begin{figure*}[!t]
    \centering
    \includegraphics[width=0.99\textwidth]{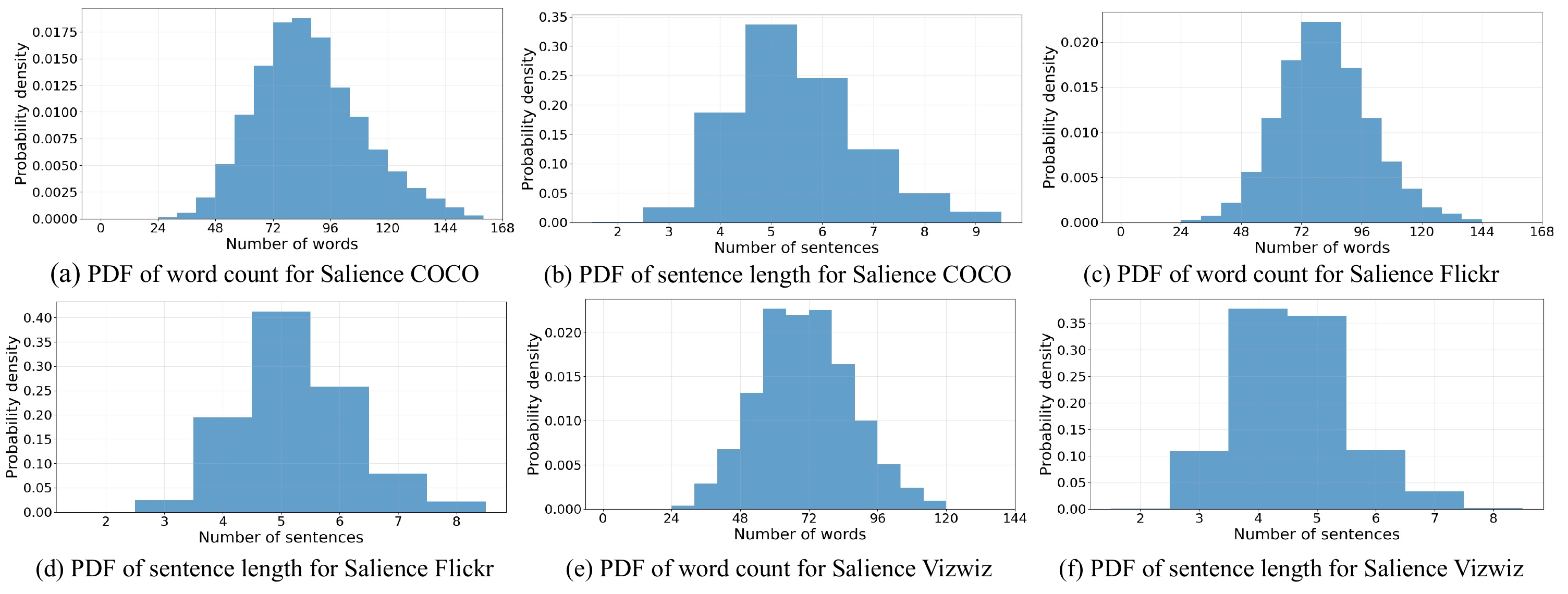}
    \caption{Statistics of word counts and sentence lengths of Salience COCO, Salience Flickr, and Salience VizWiz datasets.}
    \label{fig:stats}
\end{figure*}

\subsubsection{Salience COCO}
To address the limitations mentioned above, we repurpose the COCO dataset~\cite{lin2014microsoft} to form our Salience COCO dataset, and the repurposing process is shown in Fig.~\ref{fig:dataset}. To generate human-centered captions that effectively incorporate salience information, we redefine the ground-truth image captions. First, to capture the complexity of real-world environments, we filtered the COCO dataset \cite{lin2014microsoft} to include only images containing at least two supercategories. Next, we selected images relevant to the everyday needs of individuals with low vision, focusing on categories such as indoor objects, crossroads, and traffic signs. Each image was then processed with QAGNet\cite{deng2024advancing} to generate salience masks, where brighter regions indicate higher salience and darker regions indicate lower salience. We computed salience scores from these masks and mapped them to the original bounding boxes, ranking objects by their scores in descending order. Finally, we provided this ranked list, along with the original image, to Janus Pro 7b \cite{chen2025janus} to produce ground truth captions that capture the relative importance of each object. Eventually, our Salience COCO dataset has 6,636 training images and 1,659 test images. {To ensure reproducibility, we use a pre-trained QAGNet with a confidence threshold of 0.7, discarding low-confidence instances. For each retained object, we compute the mean saliency intensity within its bounding box and rank all objects in descending order. For caption generation, a single standardized template is applied across all three datasets, instructing Janus Pro-7B to \textit{``generate a multiple-sentence caption describing these objects exactly in the order provided.''}  This configuration, including the confidence threshold, saliency ranking procedure, and caption generation template, is kept fixed to ensure consistency. The latter two datasets, Salience Flickr and Salience VizWiz, are generated using the same configurations.}
\subsubsection{Salience Flickr}
For the Salience Flickr dataset, we generated salience-aware ground-truth captions for the Flickr30K dataset \cite{young2014image}, which does not include COCO-style annotations. We employed YOLO-World \cite{cheng2024yolo}, an annotation model based on the YOLO-V8x \cite{redmon2016you}, to automatically detect objects, extract bounding boxes, labels, supercategories, and item categories for each image. Following the same annotation process as Salience COCO, we applied a filtering procedure to maintain scenario diversity and ensure high relevance to low vision people's lives. In particular, we retain images that contain at least two distinct categories and remove those predominantly featuring outdoor sports scenes. Similar to the previously described Salience COCO ground-truth caption generation process, the correctly ordered labels and corresponding original images were then input into Janus Pro-7b\cite{chen2025janus} to generate the ground-truth captions. Ultimately, this procedure resulted in a curated dataset comprising 6,980 training images and 1,745 test images.

\subsubsection{Salience Vizwiz}
The Vizwiz dataset \cite{gurari2020captioning}, originally collected from individuals with visual impairments, offers a genuine perspective on the everyday visual challenges encountered by low vision users. To extend the utility of Vizwiz, we adopted an annotation pipeline using YOLO-World to generate COCO-style bounding boxes, labels, supercategories, and item categories for each image, as described above. We then applied a filtering procedure to retain only those images containing at least two supercategories, thereby ensuring the necessary contextual richness for effective model training. Following this, the reordered labels and corresponding original images were sent to Janus Pro-7b\cite{chen2025janus} to generate the ground-truth captions. This process ultimately yielded 779 training images and 266 test images. 

In summary, our repurposed dataset addresses the three key limitations in existing image captioning datasets for low vision assistive technologies: the absence of salience-aware information, insufficient scene complexity, and the presence of nonrepresentative contexts, such as outdoor sports, for low vision people. Through a YOLO-World ground-truth annotation pipeline and a targeted filtering process, we selected sets of training and testing images for each of the repurposed datasets that more accurately reflect the everyday visual challenges encountered by low vision users. Detailed statistics for all three repurposed datasets regarding the number of images, number of total words, number of caption sentences, average sentence numbers, vocabulary sizes, and average word counts in sentences are summarized in Table~\ref{tab:stats} and the distributions of word counts and sentence lengths for all three datasets are detailed in Fig.~\ref{fig:stats}. We also show some examples from these three datasets in Fig.~\ref{fig:data_demo}.

\begin{table}[ht]
\centering
\caption{Statistics for the three salience image captioning datasets.}
\label{tab:stats}
\resizebox{\columnwidth}{!}{%
\begin{tabular}{lccc}
\toprule
 & \textbf{Salience COCO} & \textbf{Salience Flickr} & \textbf{Salience Vizwiz} \\
\midrule
\#Images               & 8,295   & 8,725   & 1,045 \\
\#Total words        & 728,043 & 712,516 & 73,689 \\
\#Caption sentences   & 46,012  & 46,190  & 4,815 \\
Avg sentence          & 5       & 5       & 4 \\
Vocab size            & 6,657   & 6,882   & 2,854 \\
Avg words in sentence & 16      & 15      & 15 \\
\bottomrule
\end{tabular}%
}
\end{table}

{
\subsubsection{Human Verification of Annotation Quality}

To validate that our salience annotations align with the perceptual priorities of low vision users, we conducted a human verification study following established agreement evaluation protocols~\cite{cohen1960coefficient}. This evaluation examines whether our annotations order objects according to the perceived importance to real low vision users. Rather than only checking whether the same objects are present, this study also assesses whether the ranking of those objects reflects how low vision users naturally judge their importance within a scene.

To assess annotation quality, we randomly sampled 40 images from each of the three salience datasets, for a total of 120 verification images. Following Institutional Review Board approval, we recruited four low vision participants for the study. The four participants were all male, aged 38, 47, 54, and 73 years, and had a formal diagnosis of uncorrectable visual impairment with visual acuity no worse than 20/400\footnote{The subject must be at a distance of 20 feet to recognize an object that a person with normal vision can see from 400 feet away}. All were able to visually explore image details and independently complete the online survey. The verification images were divided evenly across participants, with each participant viewing 30 images. For each image, participants identified and ranked the three most salient objects based on their perceived importance. To assist this process, each image was accompanied by a GPT-5~\cite{singh2025openaigpt5card} generated description and a list of candidate objects. We then compared these human annotations with the salience rankings produced by our model on the same images.

We evaluated annotation quality using Cohen's kappa~\cite{cohen1960coefficient} between the human annotations and the model-generated salience labels on the verification images. Across all verification images, Cohen's $\kappa$ was 0.70, indicating a good agreement between the model generated salience annotations and the annotations provided by low vision participants.\footnote{{Altman's guidelines~\cite{altman1990practical} interpret Cohen's $\kappa < 0.20$ as poor, $0.21$--$0.40$ as fair, $0.41$--$0.60$ as moderate, $0.61$--$0.80$ as good, and $0.81$--$1.00$ as very good agreement.}} This result supports that our salience rankings are well aligned with the perceptual priorities of low vision users and strengthens the reliability of the annotations across the three datasets.

\section{Method}
\label{sec:method}

We propose Salience-LLaVA, a dual-branch multimodal architecture based on LLaVA~\cite{liu2023visual}, as shown in Fig.~\ref{fig:pipeline}, that enables the original LLaVA to generate salience-aware image captions. Our Salience-LLaVA consists of three components: 1) a vision backbone branch that inherits from CLIP-ViT~\cite{radford2021learning} to extract image features; 2) a salience branch built upon QAGNet~\cite{deng2024advancing} to obtain multiscale saliency features; and 3) a large language model, \ie, Vicuna~\cite{chiang2023vicuna}, to encode language instructions and output image captions. 

\subsection{Vision Backbone Branch}
We adopt CLIP-ViT as our vision encoder, as used in the original LLaVA. In particular, an input image $\mathbf{x}$ is first resized and then tokenized into a grid of nonoverlapping patches $[\mathbf{x}_1, \mathbf{x}_2, \ldots, \mathbf{x}_N]$. A ViT model embeds each patch into fixed-dimensional feature vectors, resulting in a patch-level sequence $[\mathbf{f}_1, \mathbf{f}_2, \ldots, \mathbf{f}_N] \in \mathbb{R}^{N \times d}$ where $N$ is the number of patches, and $d=768$ is the feature dimension of the ViT output. Then, an MLP projection head is applied to map these image features to match the input dimension $D = 4096$ required by Vicuna, \ie, the final output is $[\mathbf{f}_1, \mathbf{f}_2, \ldots, \mathbf{f}_N] \in \mathbb{R}^{N \times D}$.

\subsection{Multiscale Saliency Branch}
This branch begins by processing the input image \(\mathbf{x}\) through ResNet-50~\cite{he2016deep} to extract multiscale feature maps from successive convolutional layers. These feature maps, rich in both semantic and spatial information, are then refined by a pixel decoder inspired by Mask2Former~\cite{cheng2022masked} to fuse into a coherent representation that maintains a detailed spatial context. At this point, a set of $N$ learnable instance queries are introduced to act as proposals for potential salient instances. These queries are then fed into a graph neural network known as QAGNet~\cite{deng2024advancing} for saliency instance ranking. {We adopt QAGNet~\cite{deng2024advancing} as it ranks salient objects using genuine human fixation data, yielding a saliency signal that is more closely aligned with natural human visual perception. Importantly, QAGNet~\cite{deng2024advancing} does not produce fixed salience rankings; because the model is trained on human fixation patterns that naturally reflect context-dependent attention, the same object category can receive different salience scores depending on the surrounding scene. As shown in Fig.~\ref{fig:dynamic_salience}, a cup is ranked as highly salient in a dining setting but receives lower priority in a social scene where persons dominate attention, while traffic lights are ranked highest in street-crossing scenarios. This context-sensitive behavior enables our framework to adapt salience priorities across different scenes without requiring explicit task-specific rules. The overall Salience-LLaVA framework is shown in Fig.~\ref{fig:pipeline}.}

Within QAGNet~\cite{deng2024advancing}, the core architecture is organized into an input layer, multiple hidden layers, and an output layer. In the hidden layers, each QAG layer constructs a tritiered nested graph that comprises:
\begin{enumerate}
    \item Single Scale Graphs (SSGs): For each instance, three query features (obtained from different decoder layers at a fixed scale) are aggregated by averaging, and then refined using a graph neural network layer.
    \item Multiscale Graphs (MSGs): The representative nodes from the SSGs at different scales (\eg, 32, 64, 128) are combined into MSGs, capturing multiscale instance-level cues.
    \item Global Relationship Graph (GRG): MSG representative nodes from all instances are interconnected to model the global relationships among salient objects.
\end{enumerate}
The representative aggregation (RA) stage of each QAG layer aggregates these multiscale features, while the subsequent representative feedback (RF) stage propagates ranking-aware cues from the GRG back to the SSGs. This bidirectional flow refines the query representations by integrating both intrainstance multiscale details and interinstance relational information. The final output of the method is a refined feature representation $\mathbf{z} \in \mathbb{R}^{N\times D}$, also known as a saliency feature, where $N$ denotes the number of detected salient instances and $D$ is the feature dimension. This output is then passed through a linear rank head to predict the relative saliency ranking scores for each instance.

\begin{figure*}[htbp]
    \centering
    \includegraphics[width=\textwidth]{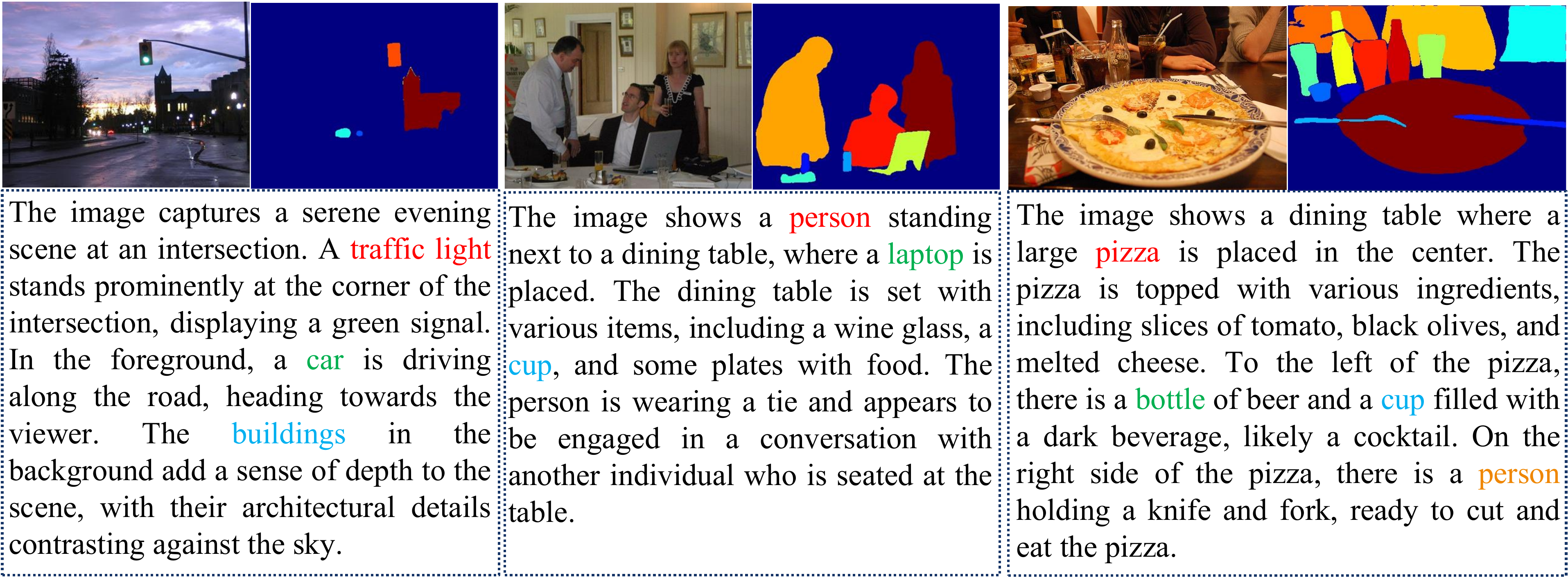}
    \caption{ {QAGNet produces context-dependent salience rankings across different scenes. Left: in a street-crossing scenario, the traffic light is ranked as the most salient object. Center: in a social setting, the person is prioritized over the cup and laptop. Right: in a dining scenario, the cup is ranked 
    above the person. These examples demonstrate that the same 
    object category (e.g., cup, person) receives different 
    salience rankings depending on the scene context.}}
    \label{fig:dynamic_salience}
\end{figure*}

\begin{figure*}[htbp]
    \centering
    \includegraphics[width=\textwidth]{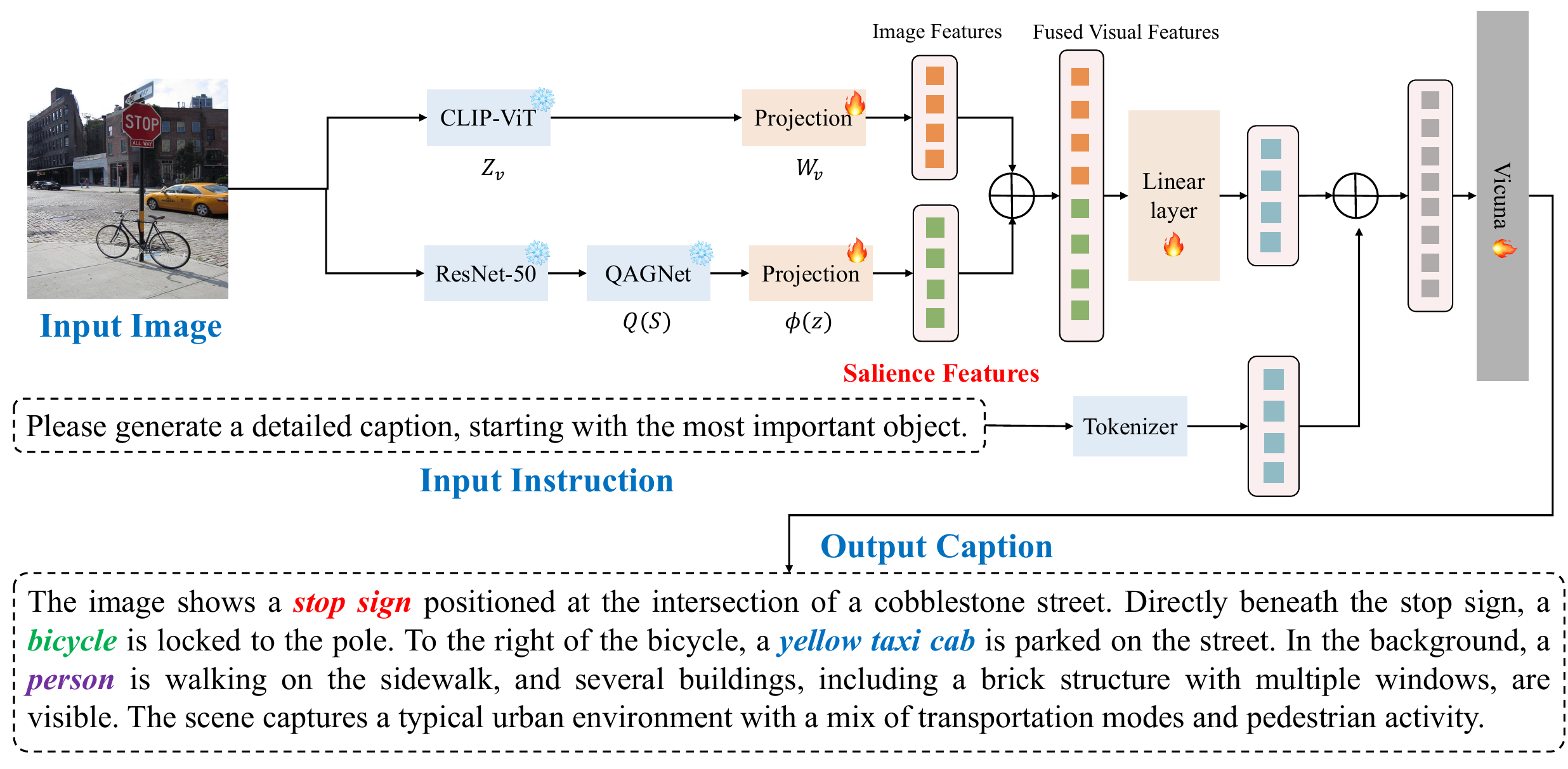}
    \caption{Overview of the Salient LLaVA framework for image captioning. The model processes an input image and an instruction, extracting visual features using CLIP-ViT and ResNet-50. A QAGNet module enhances feature extraction, followed by projection layers to generate salience features. These features are fused and processed through a linear layer, combined with tokenized instruction embeddings, and fed into Vicuna to generate the output caption.}
    \label{fig:pipeline}
\end{figure*}

\subsection{Saliency Map Construction and Fusion}
In order to capture saliency features alongside the original visual representations, we employ a projection layer that integrates both signals into a single embedding, enabling the model to leverage salient cues without disrupting the underlying visual structure. After retrieving the saliency feature $\mathbf{z}$, we pass it through a projection module consisting of a two-layer MLP defined as follows:
\begin{equation}
\label{neural_network}
\mathbf{s} = \mathbf{W_2}\,\sigma(\mathbf{W_1}\,\mathbf{z} + \mathbf{b_1}) + \mathbf{b_2}
\end{equation}
where $\mathbf{W}_1$ and $\mathbf{b}_1$ are the weight matrix and bias vector of the first linear layer, while $\mathbf{W_2}$ and $\mathbf{b_2}$ are the corresponding parameters for the second linear layer, respectively. Here, $\sigma$ denotes the ReLU function.

Next, we concatenate the projected saliency feature $\mathbf{s} \in \mathbb{R}^{N \times D_s}$ with the original feature vectors, yielding \( \mathbf{f}_{\text{fused}} \in \mathbb{R}^{N \times (D + D_s)} \). These fused features are then passed into the Vicuna for caption generation.

\subsection{Caption Generation as Response}
Language instructions, \eg, bottom left in Fig.~\ref{fig:pipeline}, are tokenized using Vicuna’s tokenizer into \(L\) tokens and embedded into a text representation \(\mathbf{e} \in \mathbb{R}^{L \times D_e}\) via the pretrained embedding layer (where \(D_e\) is the embedding dimension). Meanwhile, the vision pathway produces fused visual features \(\mathbf{f}_{\text{fused}}\). These fused features are projected to the text embedding space via a linear layer, yielding \(\hat{\mathbf{f}}_{\text{fused}} \in \mathbb{R}^{N \times D'}\), and concatenated with text representation \(\mathbf{e}\) along the sequence dimension to form a unified representation \(\mathbf{c} \in \mathbb{R}^{(L+N) \times D_e}\), which is then input into Vicuna for salience-aware caption generation.

\subsection{Loss Function}
We use the cross-entropy loss for caption generation, where for a given image-prompt pair, the target caption tokens \(y = \{y_1, \dots, y_T\}\) are generated based on the unified representation \(\mathbf{c} \in \mathbb{R}^{(L+N) \times D_e}\) as follows:

\begin{equation}
\label{loss_function}
L = -\frac{1}{T}\sum_{t=1}^{T}\log P(y_t \mid c, y_{<t})
\end{equation}

Here, \(T\) denotes the total number of tokens in the caption, and \(t\) indexes each token position. To fine-tune the VLM efficiently, we adopt low-rank adaptation (LoRA), which modifies each attention weight matrix \(W\) as \(W' = W + AB\), with \(A\) and \(B\) being low-rank trainable matrices. In addition, we only train the newly introduced projection layer for saliency integration, keeping all other model parameters frozen. This preserves the pretrained knowledge of the base model and reduces the risk of overfitting.

\section{Experiment}
\label{sec:exp}

\subsection{Evaluation}
Our evaluation framework includes both conventional captioning metrics and a newly introduced measure that assesses the salience order in which objects are identified. The conventional metrics, including BLEU, CIDEr, ROUGE, and METEOR, collectively provide complementary perspectives on the linguistic quality of the generated captions. 

To capture the salience-aware nature of our method, we introduce a ``Salience Coherent Match Index'' (SCMI) metric grounded in a longest common substring approach. Consider \(W_\text{gt}\) as the ground-truth sequence of salience object instance labels, ranked by salience, and let \(W_\text{pred}\) be the sequence of words obtained from the prediction. We define \(\texttt{LCS}(W_\text{gt}, W_\text{pred})\) as the longest subsequence common to both \(W_\text{gt}\) and \(W_\text{pred}\) without violating the salience order of the instance labels. The SCMI is defined as follows:
\begin{equation}
\label{scmi_metric}
\mathrm{SCMI}(W_\text{gt}, W_\text{pred}) 
= 
\frac{|\mathrm{LCS}(W_\text{gt}, W_\text{pred})|}{|W_\text{gt}|}
\end{equation}
This is especially relevant for low vision assistance, where describing the most salient object instances first can improve users' ability to orient themselves and navigate a scene. Notably, SCMI evaluates both the presence of key object instances and their relative ordering, rewarding predictions that preserve the salience-consistent sequence of important visual elements. {However, SCMI uses string matching and penalizes equivalent labels with different names (e.g., ``couch'' versus ``sofa''). We, therefore, report Wu-Palmer SCMI, which uses the Wu-Palmer semantic similarity~\cite{wu1994verb} to measure how closely two labels are related in meaning based on their positions in the WordNet hierarchy. In our evaluation, labels are treated as matched when their Wu-Palmer similarity exceeds 0.85, making the metric more robust to naming variations and semantically equivalent terms.}

The results are compared against three recent caption generation approaches. DeCap~\cite{li2023decap} utilizes CLIP’s multimodal embedding space through a training-free projection mechanism, which enables a text-only decoder to generate captions in a zero-shot manner. GRIT~\cite{nguyen2022grit} introduces a Transformer-based architecture that integrates grid-based and region-based features, \ie, replacing traditional CNN-based detectors with a DETR-based approach and employing a Swin Transformer backbone, to capture richer contextual details. Tag2Text~\cite{huang2024tagtext} automatically parses semantic tags from paired texts, providing strong guidance during caption generation and yielding more directed captions. {We further compare with recent works in salience-driven VLMs and low vision assistance: SCOPE~\cite{deng2025scope}, a salience-driven approach that models both saliency and semantic coverage to select the most informative visual tokens in multimodal LLMs, and WalkVLM~\cite{yuan2025walkvlm}, a vision-language model specifically designed to assist visually impaired people by generating navigation-oriented scene descriptions.}

\subsection{Results}
\label{sec:results}

\begin{table*}[ht]
\caption{Comparison of model performance on our three saliency datasets using multiple evaluation metrics. We additionally include recent state-of-the-art MLLMs evaluated with explicit salience-aware two-shot prompting. The best results for each dataset and metric are highlighted in bold.}
\label{table:results}
\centering
\resizebox{1.0\textwidth}{!}{%
\begin{tabular}{llcccccccccccc}
\toprule
 Datasets & Models & {BLEU-1} & {BLEU-2} & {BLEU-3} & {BLEU-4} & {CIDEr} & {METEOR} & {ROUGE-1} & {ROUGE-2} & {ROUGE-L} & {SCMI} & {Wu-Palmer SCMI} \\
\midrule
\multirow{10}{*}{Salience COCO} 
& DeCap \cite{li2023decap}     & 0.35 & 0.18 & 0.06 & 0.03 & 0.07 & 0.26 & 0.38 & 0.09 & 0.23 & 0.35 & 0.44 \\
& GRIT \cite{nguyen2022grit}      & 0.27 & 0.13 & 0.05 & 0.03 & 0.03 & 0.19 & 0.35 & 0.09 & 0.23 & 0.28 & 0.32 \\
& Tag2text \cite{huang2024tagtext}  & 0.27 & 0.14 & 0.06 & 0.02 & 0.05 & 0.21 & 0.36 & 0.10 & 0.24 & 0.33 & 0.38 \\
& LLaVA \cite{liu2023visual}       & 0.25 & 0.15 & 0.09 & 0.06 & 0.12 & 0.21 & 0.34 & 0.13 & 0.22 & 0.36 & 0.28 \\
& LLaVA-FT \cite{liu2023visual} & 0.46 & 0.32 & 0.22 & 0.20 & 0.40 & 0.40 & 0.55 & 0.30 & 0.35 & 0.51 & 0.58 \\
& SCOPE \cite{deng2025scope} & 0.44 & 0.27 & 0.17 & 0.10 & 0.17 & 0.33 & 0.48 & 0.19 & 0.30 & 0.53 & 0.60 \\
& WalkVLM \cite{yuan2025walkvlm} & 0.45 & 0.31 & 0.21 & 0.15 & 0.29 & 0.38 & 0.53 & 0.25 & 0.34 & 0.53 & 0.60 \\
& {Gemma4-2Shot \cite{gemma4_2026}} & {0.24} & {0.15} & {0.09} & {0.06} & {0.13} & {0.19} & {0.31} & {0.12} & {0.20} & {0.25} & {0.31} \\
& {Qwen3.6-2Shot \cite{qwen3.6-27b}} & {0.39} & {0.25} & {0.15} & {0.10} & {0.12} & {0.39} & {0.50} & {0.19} & {0.30} & {0.50} & {0.55} \\
& \textbf{Ours} 
          & \textbf{0.52} & \textbf{0.37} & \textbf{0.28} & \textbf{0.21} 
          & \textbf{0.45} & \textbf{0.43} & \textbf{0.60} & \textbf{0.33} 
          & \textbf{0.40} & \textbf{0.60} & \textbf{0.63} \\
\midrule

\multirow{10}{*}{Salience Flickr} 
& DeCap \cite{li2023decap}       & 0.35 & 0.20 & 0.16 & 0.06 & 0.06 & 0.26 & 0.37 & 0.08 & 0.21 & 0.42 & 0.51 \\
& GRIT \cite{nguyen2022grit}        & 0.28 & 0.14 & 0.06 & 0.03 & 0.03 & 0.19 & 0.35 & 0.09 & 0.23 & 0.27 & 0.29 \\
& Tag2text \cite{huang2024tagtext}    & 0.28 & 0.13 & 0.05 & 0.03 & 0.05 & 0.21 & 0.37 & 0.10 & 0.23 & 0.40 & 0.43 \\
& LLaVA \cite{liu2023visual}        & 0.30 & 0.19 & 0.12 & 0.07 & 0.13 & 0.25 & 0.39 & 0.15 & 0.25 & 0.38 & 0.27 \\
& LLaVA-FT \cite{liu2023visual}  & 0.44 & 0.32 & 0.24 & 0.20 & 0.42 & 0.37 & 0.51 & 0.28 & 0.36 & 0.73 & 0.80 \\
& SCOPE \cite{deng2025scope} & 0.41 & 0.25 & 0.15 & 0.10 & 0.14 & 0.33 & 0.46 & 0.16 & 0.29 & 0.53 & 0.59 \\
& WalkVLM \cite{yuan2025walkvlm} & 0.24 & 0.14 & 0.10 & 0.05 & 0.10 & 0.34 & 0.34 & 0.13 & 0.21 & 0.68 & 0.76 \\
& {Gemma4-2Shot \cite{gemma4_2026}} & {0.28} & {0.17} & {0.11} & {0.07} & {0.17} & {0.23} & {0.34} & {0.13} & {0.23} & {0.35} & {0.45} \\
& {Qwen3.6-2Shot \cite{qwen3.6-27b}} & {0.39} & {0.25} & {0.16} & {0.10} & {0.14} & {0.39} & {0.50} & {0.19} & {0.30} & {0.58} & {0.65} \\
& \textbf{Ours} & \textbf{0.50} & \textbf{0.35} & \textbf{0.26} & \textbf{0.20} & \textbf{0.45} & \textbf{0.42} & \textbf{0.57} & \textbf{0.31} & \textbf{0.39} & \textbf{0.79} & \textbf{0.83} \\
\midrule

\multirow{10}{*}{Salience Vizwiz} 
& DeCap \cite{li2023decap}     & 0.32 & 0.15 & 0.05 & 0.02 & 0.03 & 0.26 & 0.37 & 0.08 & 0.22 & 0.53 & 0.65 \\
& GRIT \cite{nguyen2022grit}       & 0.33 & 0.17 & 0.07 & 0.04 & 0.05 & 0.22 & 0.38 & 0.10 & 0.25 & 0.55 & 0.61 \\
& Tag2text \cite{huang2024tagtext}  & 0.26 & 0.14 & 0.06 & 0.03 & 0.05 & 0.20 & 0.37 & 0.12 & 0.25 & 0.59 & 0.62 \\
& LLaVA \cite{liu2023visual}       & 0.11 & 0.07 & 0.05 & 0.03 & 0.05 & 0.13 & 0.24 & 0.10 & 0.20 & 0.40 & 0.43 \\
& LLaVA-FT \cite{liu2023visual} & 0.35 & 0.25 & 0.18 & 0.13 & 0.31 & 0.29 & 0.43 & 0.22 & 0.30 & 0.60 & 0.49 \\
& SCOPE \cite{deng2025scope} & 0.43 & 0.27 & 0.17 & 0.11 & 0.24 & 0.33 & 0.48 & 0.20 & 0.31 & 0.65 & 0.74 \\
& WalkVLM \cite{yuan2025walkvlm} & 0.27 & 0.18 & 0.11 & 0.08 & 0.06 & 0.36 & 0.39 & 0.17 & 0.25 & 0.69 & \textbf{0.79} \\
& {Gemma4-2Shot \cite{gemma4_2026}} & {0.30} & {0.19} & {0.12} & {0.08} & {0.15} & {0.25} & {0.38} & {0.16} & {0.26} & {0.41} & {0.50} \\
& {Qwen3.6-2Shot \cite{qwen3.6-27b}} & {0.39} & {0.26} & {0.17} & {0.12} & {0.14} & {\textbf{0.40}} & {0.50} & {0.22} & {0.32} & {0.63} & {0.69} \\
& \textbf{Ours} & \textbf{0.47} & \textbf{0.32} & \textbf{0.23} & \textbf{0.17} & \textbf{0.36} & 0.38 & \textbf{0.54} & \textbf{0.27} & \textbf{0.35} & \textbf{0.71} & \textbf{0.79} \\
\bottomrule
\end{tabular}%
}
\end{table*}
In addition to DeCap~\cite{li2023decap}, GRIT~\cite{nguyen2022grit}, and Tag2text~\cite{huang2024tagtext}, we also compare our approach with the unmodified LLaVA model~\cite{liu2023visual} as well as a fine-tuned variant (LLaVA-FT) trained on our salience datasets. We further include SCOPE~\cite{deng2025scope} and WalkVLM~\cite{yuan2025walkvlm} as recent baselines representing salience-driven and low vision-oriented VLMs, respectively. {To verify whether recent MLLMs can achieve similar salience-aware captioning behavior through prompting alone, we evaluate Gemma4~\cite{gemma4_2026} and Qwen3.6~\cite{qwen3.6-27b} using two-shot examples, each consisting of an image and a ground-truth salience-aware response. We further use the same human-centered instruction as in our framework: ``Please generate a detailed human-centered caption by describing the scene from the most important object to the least important object.''} Evaluations are conducted on Salience COCO, Flickr, and VizWiz using both conventional captioning metrics (\ie, BLEU, CIDEr, METEOR, ROUGE) {and two salience-specific metrics: SCMI and Wu-Palmer SCMI}. As shown in Table~\ref{table:results}, our model {generally improves over baselines, supporting} the benefit of salience integration for generating human-centered captions. {Although Gemma4~\cite{gemma4_2026} and Qwen3.6~\cite{qwen3.6-27b} benefit from explicit salience-aware demonstrations, their performance remains below our model on most captioning and salience-ordering metrics, suggesting that prompting alone is insufficient to reliably enforce human-centered object prioritization.} On Salience COCO, while BLEU-4 shows only a small improvement over LLaVA-FT (0.21 versus 0.20), METEOR increases from 0.40 to 0.43 and ROUGE-1 from 0.55 to 0.60, with SCMI improving from 0.51 to 0.60. {SCOPE and WalkVLM achieve competitive SCMI scores of 0.53, while our model shows further improvement on SCMI and Wu-Palmer SCMI (0.60 and 0.63).} {This indicates that our method better captures salience-driven object ordering, a benefit not reflected by n-gram metrics. Overall, salience cues strengthen linguistic accuracy and perceptual alignment.}

On Salience Flickr, our method also improves over LLaVA-FT, raising BLEU-1 from 0.44 to 0.50 and CIDEr from 0.42 to 0.45, together with gains in METEOR (0.42 versus 0.37) and ROUGE-1 (0.57 versus 0.51). Although higher-order BLEU scores change only slightly, the SCMI increase from 0.73 to 0.79 highlights the stronger alignment of our captions with human salience annotations. WalkVLM achieves a notable SCMI of 0.68, though with lower conventional captioning scores. SCOPE shows moderate performance across both aspects. Our Wu-Palmer SCMI of 0.83 suggests strong semantic salience alignment on this dataset.

On the more challenging Salience VizWiz dataset, our approach demonstrates advantages. Compared to LLaVA-FT, BLEU-4 improves from 0.13 to 0.17, CIDEr from 0.31 to 0.36, and METEOR from 0.29 to 0.38, while ROUGE-1 rises from 0.43 to 0.54.  {SCMI} increases from 0.60 to 0.71, underscoring the robustness of our salience-aware design under noisy real-world conditions. {SCOPE and WalkVLM show strong SCMI performance (0.65 and 0.69), with WalkVLM matching our Wu-Palmer SCMI at 0.79, demonstrating competitive semantic salience alignment on this challenging dataset.}  These results suggest that our method generalizes well while maintaining salience alignment, as measured by SCMI and Wu-Palmer SCMI.Overall, each baseline offers distinct strengths, and incorporating salience information improves caption quality and salience alignment.

\subsection{Ablation Study}
\begin{table*}[ht]
\caption{Ablation Study on Salience Information Integration using salience COCO dataset.}
\label{table:ablation_study}
\centering
\resizebox{1.0\textwidth}{!}{
\begin{tabular}{lccccccccccc}
\toprule
\textbf{Method} & \textbf{BLEU-1} & \textbf{BLEU-2} & \textbf{BLEU-3} & \textbf{BLEU-4} & \textbf{CIDEr} & \textbf{METEOR} & \textbf{ROUGE-1} & \textbf{ROUGE-2} & \textbf{ROUGE-L} & \textbf{SCMI} & {\textbf{Wu-Palmer SCMI}} \\
\midrule
Static Merge & 0.33 & 0.21 & 0.13 & 0.08 & 0.16 & 0.25 & 0.38 & 0.15 & 0.25 & 0.41 & {0.44} \\
CNN   & 0.37 & 0.23 & 0.15 & 0.09 & 0.17 & 0.29 & 0.44 & 0.17 & 0.28 & 0.45 & {0.49} \\
\textbf{Ours}   & \textbf{0.52} & \textbf{0.37} & \textbf{0.28} & \textbf{0.21} & \textbf{0.45} & \textbf{0.43} & \textbf{0.60} & \textbf{0.33} & \textbf{0.40} & \textbf{0.60} & {\textbf{0.63}} \\
\bottomrule
\end{tabular}%
}
\end{table*}
\begin{table*}[ht]
\caption{Comparison of model performance on real-world dataset. The best results are highlighted in bold}
\label{table:real-world}
\centering
\resizebox{1.0\textwidth}{!}{%
\begin{tabular}{lccccccccccc}
\toprule
 & \textbf{BLEU-1} & \textbf{BLEU-2} & \textbf{BLEU-3} & \textbf{BLEU-4} & \textbf{CIDEr} & \textbf{METEOR} & \textbf{ROUGE-1} & \textbf{ROUGE-2} & \textbf{ROUGE-L} & \textbf{SCMI} & {\textbf{Wu-Palmer SCMI}} \\
\midrule
LLaVA \cite{liu2023visual}   & 0.14 & 0.10 & 0.06 & 0.04 & 0.03 & 0.10 & 0.16 & 0.05 & 0.12 & 0.07 & {0.20} \\
\textbf{Ours}  & \textbf{0.40} & \textbf{0.31} & \textbf{0.21} & \textbf{0.14} & \textbf{0.26} & \textbf{0.41} & \textbf{0.53} & \textbf{0.22} & \textbf{0.33} & \textbf{0.14} & {\textbf{0.56}} \\
\bottomrule
\end{tabular}%
}
\end{table*}

Based on the performance described in Section~\ref{sec:results}, we conduct ablation studies on the MLP module in the salience branch (the Projection module of the salience branch in Fig.~\ref{fig:pipeline}). In one study, we ablate the MLP module using the ``Static Merge'' strategy, which directly reshapes and concatenates the features with the original vision feature. In the other, we replace the MLP module with a CNN module. The corresponding ablation results are reported in Table~\ref{table:ablation_study}. Additional evaluations include real-world deployment in Table~\ref{table:real-world}, robustness under low-light conditions in Table~\ref{tab:lowlight}, and alternative salience networks in Table~\ref{table:ablation_swap}.

\textbf{Static Merge.} This modification results in a marked performance degradation, as shown in Table~\ref{table:ablation_study}, with the largest BLEU-1--BLEU-4 drop reaching 0.19, SCMI decreasing by 0.19, and CIDEr showing the most substantial decline of 0.29. These results indicate that the learned projection module is essential for effectively leveraging the salience features.

\textbf{CNN Substitution.} As shown in Table~\ref{table:ablation_study}, replacing the MLP with a CNN module also reduces performance; BLEU-1--BLEU-4 scores drop by up to 0.15, while CIDEr and SCMI fall by 0.28 and 0.15, respectively. These results highlight the role of the MLP module in aligning salience and vision features.

\begin{figure}[!tb]
    \centering
    \includegraphics[width=0.99\linewidth]{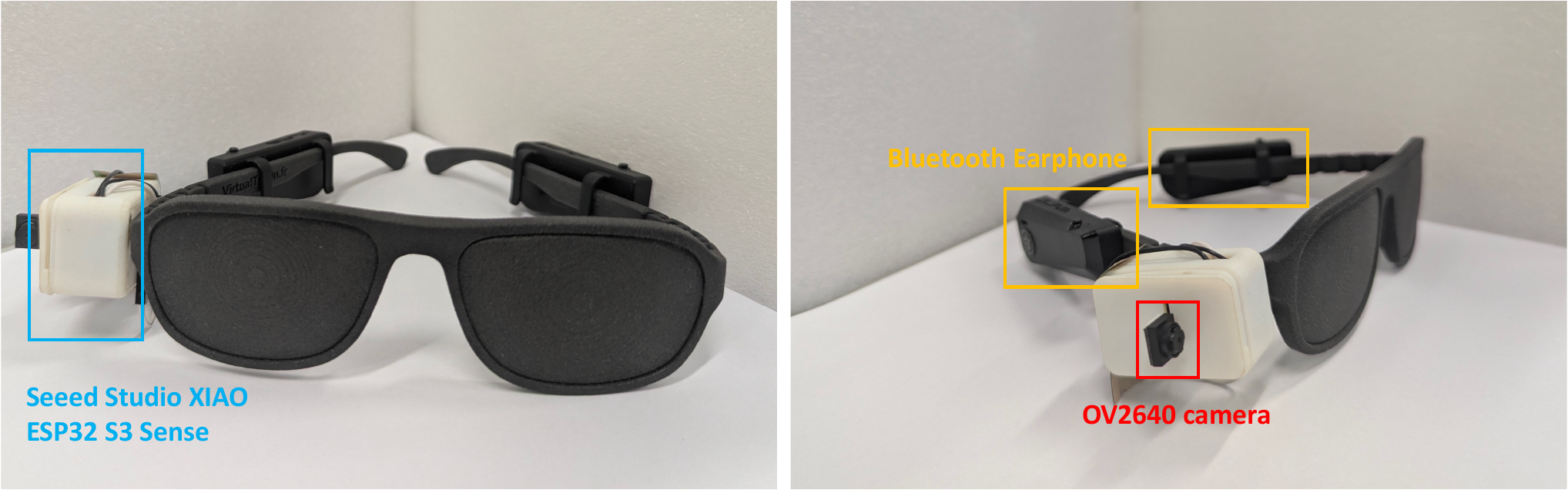}
    \caption{Smart glasses used in our real-world experiment with smart sensors, earphone and camera.}
    \label{fig:glasses}
\end{figure}

\begin{figure}[!tb]
    \centering
    \includegraphics[width=0.99\linewidth]{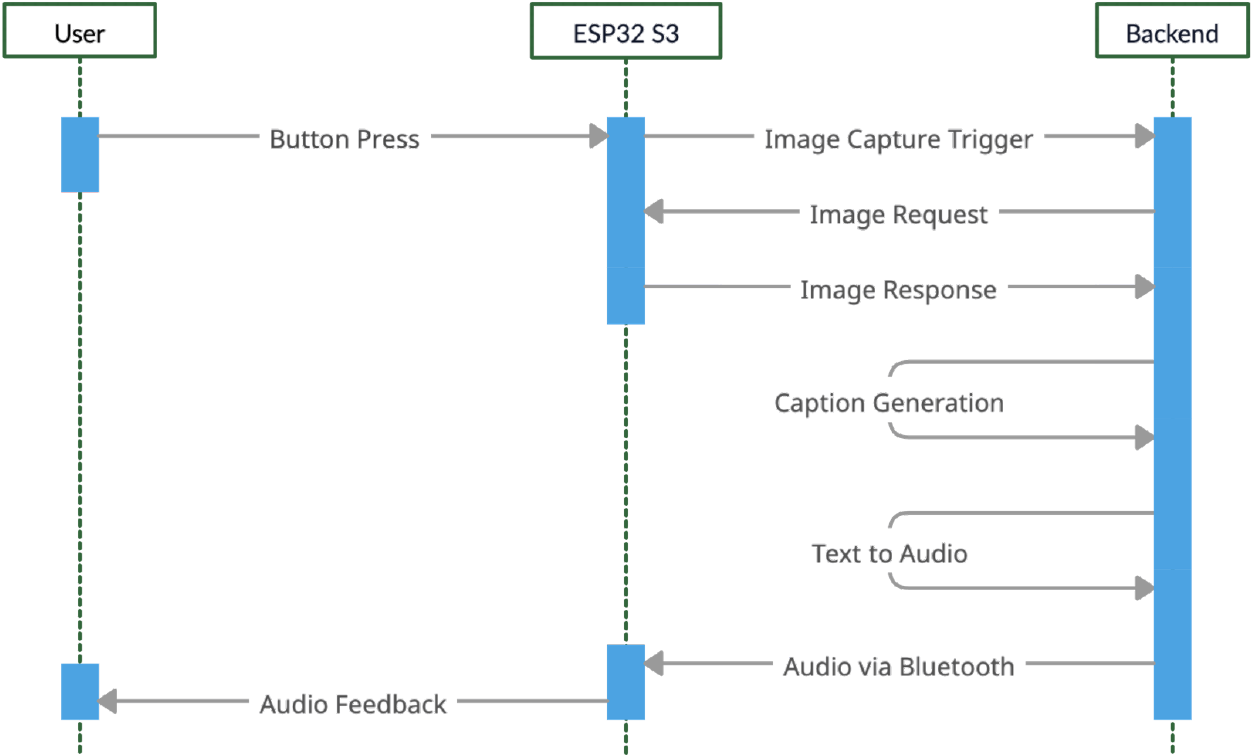}
    \caption{A sequence diagram outlines the flow of how user interacts with our captioning model deployed at the backend through ESP32 S3.}
    \label{fig:workflow}
\end{figure}

{
\subsection{Robustness Under Low-Light Conditions}

low vision users frequently encounter degraded lighting, whether due to night blindness or dimly lit environments that make scene understanding more difficult. These conditions are especially important for assistive systems because reduced visibility can affect both object recognition and the estimation of relative importance. To evaluate robustness under such conditions, we manually selected 50 low-light images from each dataset. These images are characterized by reduced brightness, weaker contrast, and limited scene visibility. We then compared model performance on these selected low-light images against performance on the full test sets to examine how the model behaves when visual quality is degraded.

As shown in Table~\ref{tab:lowlight}, performance degradation under low light is moderate overall, but the effect is more visible in salience ranking than in surface-level caption quality. On Salience COCO, BLEU-1 remains at 0.52, while SCMI drops from 0.60 to 0.53 and Wu-Palmer SCMI from 0.63 to 0.53. This suggests that the model can still generate generally reasonable descriptions, but becomes less reliable in preserving the correct ordering of important objects. A similar trend appears on Salience Flickr, where BLEU-1 stays at 0.50 while Wu-Palmer SCMI decreases from 0.83 to 0.78. On Salience VizWiz, which already contains noisy user-captured images, the degradation is larger, with Wu-Palmer SCMI falling from 0.79 to 0.63. This likely reflects the combined difficulty of poor lighting and already challenging image quality. Overall, these results suggest that reduced visibility affects QAGNet mainly in saliency estimation rather than language generation. Future improvements may, therefore, come from strengthening visual perception before ranking, such as through low-light enhancement or illumination-aware saliency estimation.

\begin{table*}[ht]
\caption{ {Robustness evaluation under low-light conditions. ``Full dataset'' denotes full test set results; ``Low-Light'' denotes the low-light subset. The best results for each dataset and metric are highlighted in bold.}}
\label{tab:lowlight}
\centering
\begingroup

\resizebox{1.0\textwidth}{!}{%
\begin{tabular}{llccccccccccc}
\toprule
\textbf{Dataset} & \textbf{Setting} & \textbf{BLEU-1} & \textbf{BLEU-2} & \textbf{BLEU-3} & \textbf{BLEU-4} & \textbf{CIDEr} & \textbf{METEOR} & \textbf{ROUGE-1} & \textbf{ROUGE-2} & \textbf{ROUGE-L} & \textbf{SCMI} & \textbf{Wu-Palmer SCMI} \\
\midrule
\multirow{2}{*}{Salience COCO}
 & Full dataset & \textbf{0.52} & \textbf{0.37} & \textbf{0.28} & \textbf{0.21} & \textbf{0.45} & \textbf{0.43} & \textbf{0.60} & \textbf{0.33} & \textbf{0.40} & \textbf{0.60} & \textbf{0.63} \\
 & Low-Light & \textbf{0.52} & 0.36 & 0.27 & 0.15 & 0.35 & 0.40 & 0.58 & 0.30 & 0.38 & 0.53 & 0.53 \\
\midrule
\multirow{2}{*}{Salience Flickr}
 & Full dataset & \textbf{0.50} & \textbf{0.35} & \textbf{0.26} & \textbf{0.20} & \textbf{0.45} & \textbf{0.42} & \textbf{0.57} & \textbf{0.31} & \textbf{0.39} & \textbf{0.79} & \textbf{0.83} \\
 & Low-Light & \textbf{0.50} & 0.34 & 0.23 & 0.16 & 0.40 & 0.39 & 0.55 & 0.24 & 0.36 & 0.75 & 0.78 \\
\midrule
\multirow{2}{*}{Salience VizWiz}
 & Full dataset & \textbf{0.47} & \textbf{0.32} & \textbf{0.23} & \textbf{0.17} & \textbf{0.36} & \textbf{0.38} & \textbf{0.54} & \textbf{0.27} & \textbf{0.35} & \textbf{0.71} & \textbf{0.79} \\
 & Low-Light & 0.43 & 0.28 & 0.20 & 0.14 & 0.32 & 0.35 & 0.51 & 0.25 & \textbf{0.35} & 0.63 & 0.63 \\
\bottomrule
\end{tabular}%
}
\endgroup
\end{table*}
\normalcolor

\subsection{Alternate Salience Network}
\begin{table*}[ht]
\caption{Salience information ablation study across three saliency datasets. The best results for each dataset and metric are highlighted in bold.}
\label{table:ablation_swap}
\centering
{
\resizebox{1.0\textwidth}{!}{%
\begin{tabular}{llcccccccccccc}
\toprule
 Datasets & Models & {BLEU-1} & {BLEU-2} & {BLEU-3} & {BLEU-4} & {CIDEr} & {METEOR} & {ROUGE-1} & {ROUGE-2} & {ROUGE-L} & {SCMI} & {Wu-Palmer SCMI} \\
\midrule
\multirow{3}{*}{Salience COCO} 
& Samba~\cite{he2025samba} & 0.51 & 0.36 & 0.27 & 0.18 & 0.41 & 0.41 & 0.59 & 0.31 & 0.38 & 0.57 & 0.58 \\
& DSGNN~\cite{wu2024domain} & 0.51 & 0.35 & 0.25 & 0.19 & 0.42 & 0.41 & 0.59 & 0.31 & 0.38 & 0.57 & 0.59 \\
& \textbf{Ours} & \textbf{0.52} & \textbf{0.37} & \textbf{0.28} & \textbf{0.21} & \textbf{0.45} & \textbf{0.43} & \textbf{0.60} & \textbf{0.33} & \textbf{0.40} & \textbf{0.60} & \textbf{0.63} \\
\midrule
\multirow{3}{*}{Salience Flickr} 
& Samba~\cite{he2025samba} & \textbf{0.52} & \textbf{0.38} & \textbf{0.29} & \textbf{0.23} & 0.40 & \textbf{0.43} & \textbf{0.59} & \textbf{0.31} & \textbf{0.41} & 0.69 & 0.77 \\
& DSGNN~\cite{wu2024domain} & \textbf{0.52} & \textbf{0.38} & \textbf{0.29} & 0.22 & 0.41 & 0.42 & \textbf{0.59} & \textbf{0.31} & \textbf{0.41} & 0.70 & 0.76 \\
& \textbf{Ours} & 0.50 & 0.35 & 0.26 & 0.20 & \textbf{0.45} & 0.42 & 0.57 & \textbf{0.31} & 0.39 & \textbf{0.79} & \textbf{0.83} \\
\midrule
\multirow{3}{*}{Salience Vizwiz} 
& Samba~\cite{he2025samba} & \textbf{0.47} & 0.31 & 0.21 & \textbf{0.17} & 0.31 & 0.32 & 0.52 & \textbf{0.28} & 0.31 & 0.66 & 0.73 \\
& DSGNN~\cite{wu2024domain} & 0.46 & 0.30 & 0.19 & 0.15 & 0.32 & 0.30 & \textbf{0.54} & 0.25 & 0.33 & 0.68 & 0.74 \\
& \textbf{Ours} & \textbf{0.47} & \textbf{0.32} & \textbf{0.23} & \textbf{0.17} & \textbf{0.36} & \textbf{0.38} & \textbf{0.54} & 0.27 & \textbf{0.35} & \textbf{0.71} & \textbf{0.79} \\
\bottomrule
\end{tabular}%
}
}
\end{table*}

{To further examine whether our framework depends on a salience extractor, we replace QAGNet with two recent salience models, Samba~\cite{he2025samba} and DSGNN~\cite{wu2024domain}, while keeping the remaining Salience-LLaVA architecture unchanged. As shown in Table~\ref{table:ablation_swap}, Samba~\cite{he2025samba} and DSGNN~\cite{wu2024domain} achieve competitive performance. For some conventional captioning metrics such as BLEU, ROUGE, and METEOR, these alternative salience models obtain comparable results, with a few scores slightly higher.

On the contrary, our model consistently performs better on SCMI and Wu-Palmer SCMI, the two metrics most directly tied to human-centered salience ordering. This gap indicates that while alternative salience models can provide useful visual cues for caption generation, they tend to order objects less consistently with human priorities, causing less important elements to sometimes appear before more critical ones. Our method produces object orderings that more closely follow human salience annotations, resulting in higher SCMI and Wu-Palmer SCMI scores.}

\subsection{Real-World Experiment}
We design and implement a wearable device, as shown in Fig.~\ref{fig:glasses}, to assist visually impaired users by providing almost real-time scene analysis through auditory feedback. At its core is the Seeed Studio XIAO ESP32 S3 Sense, a compact, thumb-sized development board that integrates an OV2640 camera sensor along with Wi-Fi and Bluetooth connectivity. This powerful yet small microcontroller captures high-resolution images and facilitates efficient wireless communication, suited for integration into a wearable platform. The interaction flow between the user, wearable device, and backend captioning model is illustrated in Fig.~\ref{fig:workflow}.

{
In the real-world experiment, we used the same pipeline as the Salience datasets: YOLO-World extracts COCO-style annotations, and Janus Pro-7b generates ground-truth captions. As shown in Table~\ref{table:real-world}, our method outperforms the original LLaVA on BLEU, CIDEr, METEOR, and ROUGE, while SCMI remains relatively low. This drop is primarily attributable to the hardware constraints of the edge device used for image acquisition. Because of its compact, portable form factor, the onboard camera operates at a limited resolution, which reduces the level of visual detail captured in each frame and diminishes the clarity of objects within the scene. Under these conditions, the model struggles to distinguish fine-grained object categories, and the resulting detections are often less specific, occasionally collapsing visually similar classes such as different vehicle or furniture types into a single broader label. Consequently, the model tends to produce less precise classifications, which limits its ability to generate exact object labels and yields lower SCMI scores than those observed on benchmark datasets with higher image resolutions.}

 This resolution limitation directly explains the larger Wu-Palmer improvement on the real-world dataset. Because images captured by the edge device have lower visual clarity, the model often predicts labels that are semantically correct but not exact lexical matches to the reference annotations. For example, the model may output a broader but still appropriate term such as vehicle'' instead of a more specific label such as car,'' or use a related category name that reflects the visible content at reduced fidelity. Under exact matching, these predictions are penalized even when they preserve the underlying meaning. By applying Wu-Palmer similarity with a threshold of 0.85, the evaluation can account for semantic relatedness in WordNet and better reflect whether the model identified the correct concept. This leads to a substantial increase in SCMI, from 0.14 to 0.56, showing that much of the apparent error comes from naming variations rather than a failure to capture the salient object. {To further assess real-world usability, we conducted a pilot study with two participants in real-world scenes, including indoor areas and street-crossing environments. For each scene, participants viewed the captured image and identified the three most important objects for understanding the environment in the order of importance. We then compared the objects selected by participants with the top three salient objects conveyed through the Salience-LLaVA audio output. The Wu-Palmer SCMI was 0.64, and Cohen's kappa~\cite{cohen1960coefficient} was 0.62, indicating a good agreement between participant selections and model outputs. These results provide preliminary evidence that the proposed system can preserve human-aligned salience cues during real-world wearable deployment and deliver useful audio feedback in practical environments.} Despite these challenges, the improvements across most metrics indicate the robustness of our approach in unconstrained, real-world scenarios.

\section{Conclusion}

{In this work, we present Salience-LLaVA, a salience-driven VLM, along with three salience-aware datasets and a dedicated evaluation metric (SCMI), to enhance accessibility for individuals with low vision.} By integrating saliency cues and reordering visual objects based on real-world prioritization, Salience-LLaVA generates descriptions that better align with human attention and intent. Evaluations on three benchmark datasets demonstrate consistent improvements over existing methods across standard captioning metrics, highlighting the model's ability to produce contextually relevant and perceptually meaningful outputs. We further validate its practical effectiveness through deployment on a customized wearable device, showcasing its real-time applicability in everyday assistive scenarios.

{In terms of limitations, the current model remains computationally heavy, requiring server-based inference and limiting fully mobile use when network access is unreliable or unavailable. Beyond low vision assistance, the framework could extend to autonomous navigation, search and rescue, and industrial inspection, where salient information must also be prioritized. Future work will improve edge efficiency and evaluate more diverse real-world scenarios, including outdoor navigation, public transit, and crowded urban environments, to better reflect everyday conditions faced by low vision users. }

\section*{Acknowledgment}
The authors thank Dr. William Seiple and Fernanda Garcia-Pi\~{n}a for questionnaire feedback and participant recruitment; the study was approved by the Lighthouse Guild's IRB.

\bibliographystyle{IEEEtran}
\bibliography{egbib}

\end{document}